\documentclass[10pt,twocolumn,letterpaper]{article}

\usepackage{cvpr}
\usepackage{times}
\usepackage{epsfig}
\usepackage{graphicx}
\usepackage{amsmath}
\usepackage{amssymb}
\usepackage{float}
\usepackage{multirow}
\usepackage{bm}
\usepackage{wrapfig}
\usepackage{authblk}

\newcommand{\bS}{{\bf S}}
\newcommand{\bT}{{\bf T}}
\newcommand{\bB}{{\bf B}}
\newcommand{\balpha}{{\bm \alpha}}
\newcommand{\bbeta}{{\bm \beta}}
\newcommand{\bdelta}{{\bm \delta}}
\newcommand{\bgamma}{{\bm \gamma}}
\newcommand{\bc}{{\bm c}}
\newcommand{\bp}{{\bf p}}
\newcommand{\bR}{{\bf R}}
\newcommand{\bt}{{\bf t}}
\newcommand{\bx}{{\bf x}}

\DeclareMathOperator*{\argmax}{arg\,max}

\usepackage[pagebackref=true,breaklinks=true,letterpaper=true,colorlinks,bookmarks=false]{hyperref}

\cvprfinalcopy % *** Uncomment this line for the final submission

\def\cvprPaperID{22} % *** Enter the CVPR Paper ID here

\ifcvprfinal\fi
\begin{document}
	
%%%%%%%%% TITLE
\title{
	\vspace{-6pt}
	Accurate 3D Face Reconstruction with Weakly-Supervised Learning:\\ From Single Image to Image Set\\
	\vspace{-4pt}
}
\author{
Yu Deng\thanks{This work was done when Yu Deng was an intern at MSRA.}\,\,$^{1,2}$ \quad Jiaolong Yang$^{2}$ \quad Sicheng Xu$^{3,2}$ \quad Dong Chen$^{2}$ \quad Yunde Jia$^{3}$ \quad Xin Tong$^{2}$ \\
$^1${Tsinghua University} \quad  $^2${Microsoft Research Asia} \quad $^3${Beijing Institute of Technology}\\

{\tt\small \{t-yudeng,jiaoyan,doch,xtong\}@microsoft.com, \{xusicheng,jiayunde\}@bit.edu.cn}

}

\maketitle

%\thispagestyle{empty}

%%%%%%%%% ABSTRACT
\begin{abstract}
	Recently, deep learning based 3D face reconstruction methods have shown promising results in both quality and efficiency.
	However, training deep neural networks typically requires a large volume of data, whereas face images with ground-truth 3D face shapes are scarce. 
	In this paper, we propose a novel deep 3D face reconstruction approach that 1) leverages a robust, hybrid loss function for weakly-supervised learning which takes into account both low-level and perception-level information for supervision, and 2) performs multi-image face reconstruction by exploiting complementary information from different images for shape aggregation. Our method is fast, accurate, and robust to occlusion and large pose. We provide comprehensive experiments on three datasets, systematically comparing our method with fifteen recent methods and demonstrating its state-of-the-art performance. Code available at  \emph{\url{https://github.com/Microsoft/Deep3DFaceReconstruction}}
\end{abstract}

%%%%%%%%% BODY TEXT
\section{Introduction}

Faithfully recovering the 3D shapes of human faces from unconstrained 2D images is a challenging task and has numerous applications such as face recognition~\cite{blanz2003face,tran2017regressing,zulqarnain2018learning}, face media manipulation~\cite{blanz1999morphable,thies2016face2face}, and face animation~\cite{cao20133d,hu2017avatar}. 
Recently, there is a surge of interest in 3D face reconstruction from a single image using deep Convolutional Neutral Networks (CNN) in lieu of the complex and costly optimization used by traditional methods~\cite{richardson20163d,dou2017end,richardson2017learning,tran2017regressing,jackson2017large,sela2017unrestricted,tewari2017mofa,tewari2018self,tran2018nonlinear,sengupta2018sfsnet,feng2018joint,guo2018cnn}. Since ground truth 3D face data is scarce, many previous approaches resort to synthetic data or using 3D shapes fitted by traditional methods as surrogate shape labels~\cite{richardson20163d,zhu2016face,sela2017unrestricted,liu2018disentangling,feng2018joint,guo2018cnn}. However, their accuracy may be jeopardized by the domain gap issue or the imperfect training labels.

To circumvent these issues, methods have been proposed to train networks without shape labels in an unsupervised or weakly-supervised fashion and promising results have been obtained~\cite{tewari2017mofa,tewari2018self,tran2018nonlinear,sengupta2018sfsnet,genova2018unsupervised}. The crux of unsupervised learning is a differentiable image formation procedure which renders a face image with the network predictions, and the supervision signal stems from the discrepancy between the input image and the rendered counterpart. For example, Tewari \etal~\cite{tewari2017mofa} and Sengupta \etal~\cite{sengupta2018sfsnet} use pixel-wise photometric difference as training loss.
To improve robustness and expressiveness, Tewari \etal~\cite{tewari2018self} proposed a two-step reconstruction scheme where the second step produces a shape and texture correction with a neural network. Genova \etal~\cite{genova2018unsupervised} proposed to measure face image discrepancy on the perception level by using the distances between deep features extracted from a face recognition network.

Our goal in this paper is to obtain accurate 3D face reconstruction with weakly-supervised learning. We identified that using low-level information of pixel-wise color alone may suffer from local minimum issue where low error can be obtained with unsatisfactory face shapes. 
On the other hand, using only perceptual loss also lead to sub-optimal results since it ignores the pixel-wise consistency with raw image signal. In light of this, we propose a hybrid-level loss function which integrates both of them, giving rise to accurate results. We also propose a novel skin color based photometric error attention strategy, granting our method further robustness to occlusion and other challenging appearance variations such as beard and heavy make-up.
We train an off-the-shelf deep CNN to predict 3D Morphable Model (3DMM)~\cite{blanz1999morphable} coefficients, and accurate reconstruction is achieved on multiple datasets~\cite{bagdanov2011florence,cao2014facewarehouse,yin20063d}.

With a strong CNN model for single-image 3D face reconstruction, we take a further step and consider the problem of CNN-based face reconstruction aggregation with a set of images.
Given multiple face images of a subject (\eg, from a personal album) captured in the wild under disparate conditions, it is natural to leverage all the images to build a better 3D face shape.
To apply the deep neural networks on arbitrary number of images, one solution would be aggregating the single-image reconstruction results, and perhaps the simplest strategy is naively averaging the recovered shapes. However, such a native strategy did not consider the quality of the input images (\eg, if some samples contain severe occlusion). Nor does it take full advantage of pose differences to improve the shape prediction.

In this paper, we propose to learn 3D face aggregation from multiple images, also in an unsupervised fashion. We train a simple auxiliary network to produce ``confidence scores" of the regressed identity-bearing 3D model coefficients, and obtain final identity coefficients via confidence-based aggregation. Despite no explicit confidence label is used, our method automatically learns to favor high-quality (especially high-visibility) photos. Moreover, it can exploit pose difference to better fuse the complementary information, learning to more accurate 3D shapes.

To summarize, this paper makes the following two main contributions:
\vspace{-6pt}
\begin{itemize}
	\item We propose a CNN-based single-image face reconstruction method which exploits hybrid-level image information for weakly-supervised learning. Our loss consists of a robustified image-level loss and a perception-level loss. We demonstrate the benefit of combing them, and show the state-of-the-art accuracy of our method on multiple datasets \cite{bagdanov2011florence,cao2014facewarehouse,yin20063d}, significantly outperforming previous methods trained in a fully supervised fashion 
	\cite{sela2017unrestricted,feng2018joint,tran2017regressing}. Moreover, we show that with a low-dimensional 3DMM subspace, we are still able to outperform prior art with ``unrestricted" 3D representations \cite{sela2017unrestricted,tran2018nonlinear,tewari2018self,feng2018joint} by an appreciable margin.
	
	\item We propose a novel shape confidence learning scheme for multi-image face reconstruction aggregation. Our confidence prediction subnet is also trained in a weakly-supervised fashion without ground-truth label. We show that our method clearly outperforms naive aggregation (\eg, shape averaging) and some heuristic strategies~\cite{piotraschke2016automated}. To our knowledge, this is the first attempt towards CNN-based 3D face reconstruction and aggregation from an unconstrained image set.
\end{itemize}

%-------------------------------------------------------------------------
\section{Related Work}

3D face reconstruction has been a longstanding task in computer vision and computer graphics.
In the literature, 3D Morphable Models (3DMM)~\cite{blanz1999morphable,paysan20093d,booth20163d} have played a paramount role for 3D face modelling. With a 3DMM, reconstruction can be performed by an analysis-by-synthesis scheme using image intensity~\cite{blanz1999morphable} and other features such as edges~\cite{romdhani2005estimating}.
More recently, model fitting using facial landmarks has gained much popularity with the growth of face alignment techniques~\cite{blanz2004statistical,zhu2015high,hassner2015effective,bas2016fitting}. However, sparse landmarks cannot well capture the dense facial geometry so these methods are poor at preserving facial fidelity. Beyond 3DMM, another popular 3D face model is the multilinear tensor model~\cite{vlasic2005face,cao20133d,cao2014displaced,saito2016real}.
A few model-free single-image reconstruction methods have been proposed~\cite{hassner2006example,kemelmacher2011face,hassner2013viewing}; most of them require some reference 3D face shapes. For example, \cite{hassner2006example,hassner2013viewing} estimate image depth by building correspondences between the input image and one or a set of reference 3D faces. In \cite{kemelmacher2011face}, a shape-from-shading approach is proposed with a reference 3D face as prior.

The aforementioned approaches usually involve costly optimization process to recover a high quality 3D face. Recently, numerous methods are proposed which employ CNNs to achieve efficient face reconstruction.
Some of them apply CNNs to regress 3DMM coefficients directly~\cite{richardson20163d,dou2017end,bas20173d,tewari2017mofa,tran2017regressing,genova2018unsupervised}, some use multi-step schemes to add correction or details onto coarse 3DMM predictions~\cite{richardson2017learning,tewari2018self,tran2018extreme,guo2018cnn}, while others advocate direct model-free reconstruction~\cite{sela2017unrestricted,tran2018nonlinear,sengupta2018sfsnet,feng2018joint}.

For all these CNN-based methods, one great hurdle is the lack of training data. To alleviate the problem, many methods resort to synthetic data or using 3D shapes fitted by traditional methods as surrogate training labels~\cite{richardson20163d,zhu2016face,sela2017unrestricted,liu2018disentangling,feng2018joint,guo2018cnn}. Others have attempted unsupervised or weakly-supervised training ~\cite{tewari2017mofa,tewari2018self,tran2018nonlinear,sengupta2018sfsnet,genova2018unsupervised}. Our method is also based on weakly-supervised learning, for which our findings in this paper are threefold: {\textbf{1)}} the loss function is important for weakly-supervised learning and both low-level and perception-level information should be leveraged; {\textbf{2)}} the results obtained with weak supervision can be significantly better than those trained with synthetic data or pseudo ground truth shapes, and {\textbf{3)}} somewhat surprisingly, the results confined in the low-dimensional 3DMM subspace can still be much better than state-of-the-art results with ``unrestricted" representations.

We also studied the problem of reconstruction aggregation from multiple images. One related work is \cite{piotraschke2016automated} which investigated the reconstruction quality measurement closest to human ratings and used it to fuse the reconstructions obtained with 3DMM fitting. We however show that their method is deficient in our case.
%for quality (confidence) measurement.
Our method is also related to traditional methods working on unconstrained photo collections~\cite{kemelmacher2011face,suwajanakorn2014total,roth2015unconstrained,roth2016adaptive}.
While excellent results have been obtained by these methods, they typically consist of multiple steps such as face frontalization, photometric stereo, and local normal refinement. The whole pipeline is complex and may break down under severe occlusion and extreme pose.
Our goal in this paper is not to replace these traditional methods, but to study the shape aggregation problem (similar to \cite{piotraschke2016automated}) with a CNN and 
provide an extremely fast and robust alternative learned end-to-end.

%------------------------------------------------------------------------

\begin{figure*}[t!]
	\vspace{-3pt}
	\centering
	\includegraphics[width=0.97\textwidth]{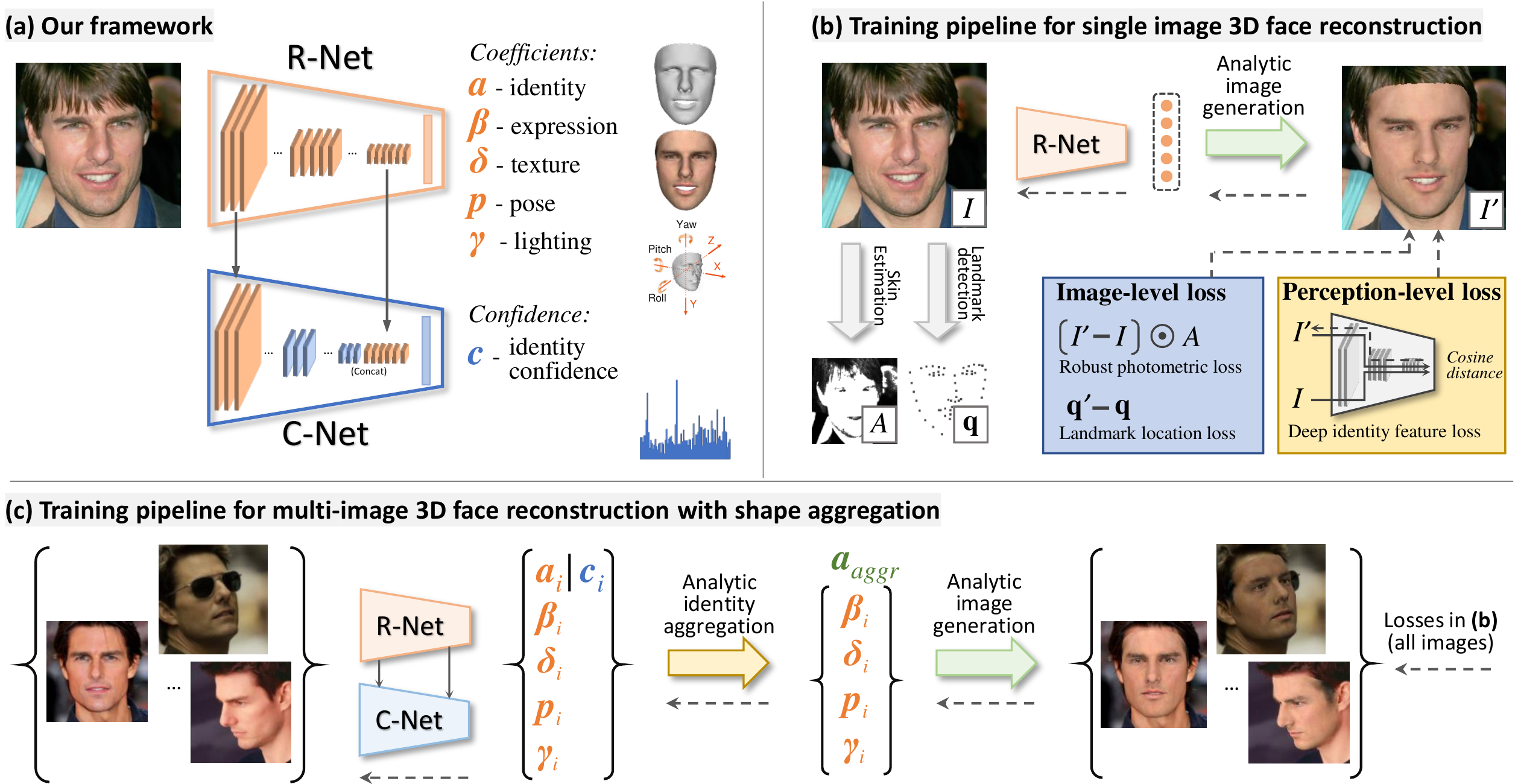}
	\caption{Overview of our approach. \textbf{(a)} The framework of our method, which consists of a reconstruction network for end-to-end single image 3D reconstruction and a confidence measurement subnet designed for multi-image based reconstruction. \textbf{(b)} The training pipeline for single images with our proposed hybrid-level loss functions. Our method does not require any ground-truth 3D shapes for training. It only leverages some weak supervision signals such as facial landmarks, skin mask and a pre-trained face recognition CNN. \textbf{(c)} The training pipeline for multi-image based reconstruction. Our confidence subnet learns to measure the reconstruction confidence for aggregation with out any explicit label. The dashed arrows denote error backpropagration for network training.}\label{fig:overview}
\end{figure*}

\section{Preliminaries: Models and Outputs}

As shown in Fig.~\ref{fig:overview} (a), we use a CNN to regress coefficients of a 3DMM face model. For unsupervised/weakly-supervised training~\cite{tewari2017mofa,tewari2018self}, we also regress the illumination and face pose to enable analytic image regeneration. We detail our models and CNN outputs as follows.

\vspace{6pt}
\noindent\textbf{3D Face Model.~}
With a 3DMM, the face shape $\bS$ and the texture $\bT$ can be represented by an affine model:
\begin{equation}
\begin{split}
\bS &= \bS(\balpha,\bbeta) = \bar{\bS} + \bB_{id}\balpha + \bB_{exp}\bbeta \\
\bT &= \bT(\bdelta) = \bar{\bT} + \bB_t\bdelta
\end{split}\label{equation:MM}
\end{equation}
where $\bar{\bS}$ and $\bar{\bT}$ are the average face shape and texture; $\bB_{id}$,  $\bB_{exp}$, and $\bB_t$ are the PCA bases of identity, expression, and texture respectively, which are all scaled with standard deviations; $\balpha$, $\bbeta$, and $\bdelta$ are the corresponding coefficient vectors for generating a 3D face. We adopt the popular 2009 Basel Face Model~\cite{paysan20093d} for $\bar{\bS}$, $\bB_{id}$, $\bar{\bT}$, and $\bB_t$, and use the expression bases $\bB_{exp}$ of \cite{guo2018cnn} which are built from FaceWarehouse~\cite{cao2014facewarehouse}. A subset of the bases is selected, resulting in $\balpha\in \mathbb{R}^{80}$, $\bbeta \in \mathbb{R}^{64}$ and $\bdelta \in \mathbb{R}^{80}$. We exclude the ear and neck region, and our final model contains 36K vertices.

\vspace{6pt}
\noindent\textbf{Illumination Model.~}
We assume a Lambertian surface for face and approximate the scene illumination with Spherical Harmonics (SH)~\cite{ramamoorthi2001efficient,ramamoorthi2001signal}. The radiosity of a vertex $\mathbf{s}_i$ with surface normal $\mathbf{n}_i$ and skin texture $\mathbf{t}_i$ can then be computed as $\mathbf{C}(\mathbf{n}_i,\mathbf{t}_i | \bgamma) = \mathbf{t}_i\!\cdot\! \sum_{b = 1}^{B^2} \gamma_b \Phi_b(\mathbf{n}_i)$
where $\Phi_b\!:\!\mathbb{R}^{3}\!\to\!\mathbb{R}$ are SH basis functions and $\gamma_b$ are the corresponding SH coefficients. We choose $B=3$ bands following \cite{tewari2017mofa,tewari2018self} and assume monochromatic lights such that $\bgamma\in\mathbb{R}^9$.

\vspace{6pt}
\noindent\textbf{Camera Model.~} We use the perspective camera model with an empirically-selected focal length for the 3D-2D projection geometry. The 3D face pose $\bp$ is represented by rotation $\bR \in \mathrm{SO}(3)$ and translation $\bt\in \mathbb{R}^3$.

In summary, the unknowns to be predicted can be represented by a vector
$\bx=(\balpha,\bbeta,\bdelta,\bgamma,\bp)\in\mathbb{R}^{239}$.
In this paper, we use a ResNet-50 network~\cite{he2016deep} to regress these coefficients by modifying the last fully-connected layer to 239 neurons. For brevity, we denote this modified ResNet-50 network for single image reconstruction as R-Net. We present how we train it in the following section.

\section{Hybrid-level Weak-supervision for Single-Image Reconstruction}\label{sec:hyrid}

Given a training RGB image $I$, we use R-Net to regress a coefficient vector $\bx$, with which a reconstructed image $I'$ can be analytically generated with some simple, differentiable math derivations
%(details are deferred to the \emph{suppl. material})
. Some examples of $I'$ can be found in Fig.~\ref{fig:overview}. Our R-Net is trained without any ground truth labels, but via evaluating a hybrid-level loss on $I'$ and backpropagate it.

\subsection{Image-Level Losses}

We first introduce our loss functions on low-level information including per-pixel color and sparse 2D landmarks.

\begin{figure}[t!]
	\includegraphics[width=1.0\columnwidth]{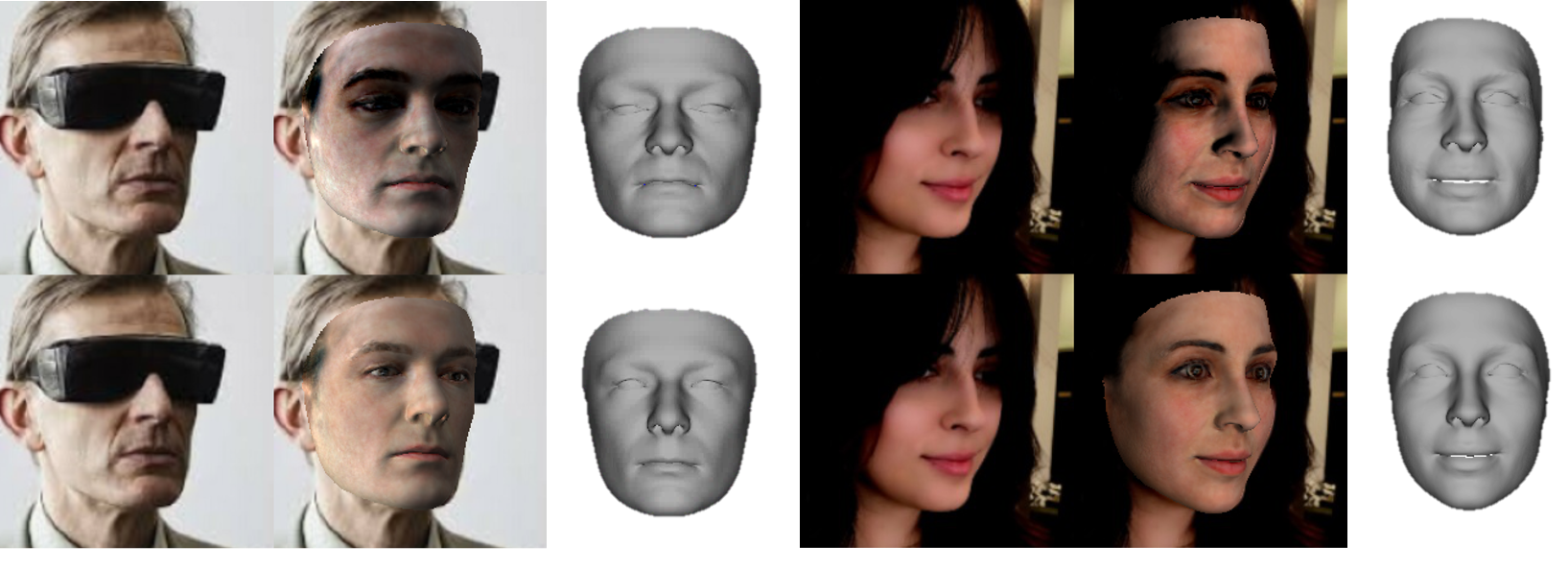}
	\caption{Comparison of the results without (\emph{top row}) and with (\emph{bottom row}) using our skin attention mask for training.}\label{fig:mask}
\end{figure}

\subsubsection{Robust Photometric Loss}
First, it is straightforward to measure the dense photometric discrepancy between the raw image and the reconstructed one \cite{blanz1999morphable,thies2016face2face,tewari2017mofa,tewari2018self}. In this paper, we propose a robust, skin-aware photometric loss instead of a naive one, defined as:
\begin{equation}
\begin{split}
L_{photo}(\bx) = \frac{\sum_{i \in \mathcal{M}}A_i \cdot \| I_i - I_i'(\bx) \|_{2}}{\sum_{i \in \mathcal{M}}A_i}\label{equation:photometric}
\end{split}
\end{equation}
where $i$ denotes pixel index, $\mathcal{M}$ is reprojected face region which can be readily obtained, $\|\!\cdot\!\|$ denotes the $l_2$ norm, and $A$ is a skin color based attention mask for the training image which is described as follows.

\vspace{6pt}
\noindent\textbf{Skin Attention.~} To gain robustness to occlusions and other challenging appearance variations such as beard and heavy make-up, we compute a skin-color probability $P_i$ for each pixel. We train a naive Bayes classifier with Gaussian Mixture Models on a skin image dataset from \cite{jones2002statistical}
%(details can be found in the \emph{suppl. material})
. For each pixel $i$, we set
$A_i =$ {\footnotesize $\begin{cases}
	1, &\!\!\!\text{if~~} P_i > 0.5 \vspace{-3pt}\\ 
	P_i, &\!\!\!\text{otherwise}
	\end{cases}$.}
We find that such a simple skin-aware loss function works remarkably well in practice without the need for a face segmentation method~\cite{saito2016real}. Figure~\ref{fig:mask} illustrates the benefit of using our skin attention mask.

It is also worth mentioning that our loss in Eq.~\ref{equation:photometric} integrates over 2D image pixels as opposed to 3D shape vertices in \cite{tewari2017mofa,tewari2018self}. It enables us to easily identify self-occlusion via z-buffering thus our trained model can handle large poses.

\subsubsection{Landmark Loss}\label{section:lm}
We also use landmark locations on the 2D image domain as weak supervision to train the network. We run the state-of-the art 3D face alignment method of \cite{bulat2017far} to detect 68 landmarks $\{\mathbf{q}_n\}$ of the training images. During training, we project the 3D landmark vertices of our reconstructed shape onto the image obtaining $\{\mathbf{q}'_n\}$, and compute the loss as:
\begin{equation}
L_{lan}(\bx) =  \frac{1}{N}\sum_{n=1}^{N}\omega_n \| \mathbf{q}_n - \mathbf{q}'_n(\bx) \|^2
\end{equation}
where $\omega_n$ is the landmark weight which we experimentally set to $20$ for inner mouth and nose points and others to $1$.

\begin{figure}[t!]
	\includegraphics[width=1.0\columnwidth]{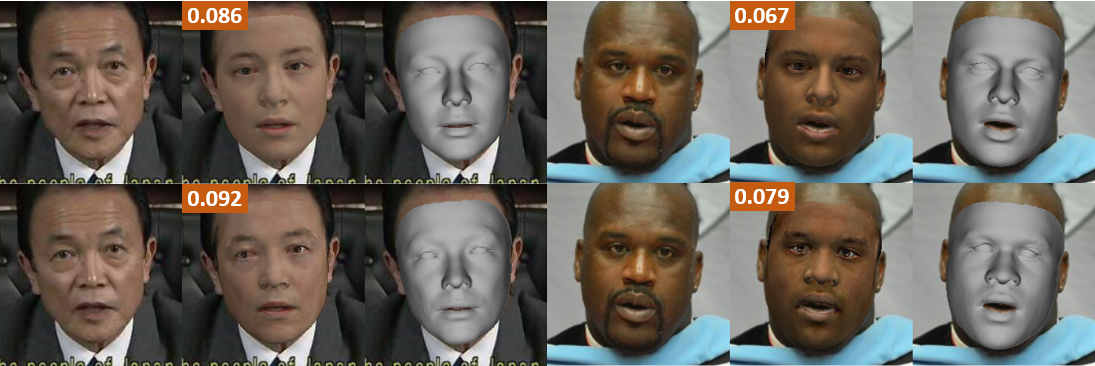}
	\vspace{-10pt}
	\caption{Comparison of the results with only image-level losses (\emph{top row}) and with both image-level and perceptual losses (\emph{bottom row}) for training. The numbers are the evaluated photometric errors. A lower photometric error does not guarantee a better shape.}\label{fig:id}
\end{figure}

\subsection{Perception-Level Loss}
While using the low-level information to measure image discrepancy can generally yields decent results~\cite{blanz1999morphable,thies2016face2face,tewari2017mofa,tewari2018self}, we find using them alone can lead to local minimum issue for CNN-based 3D face reconstruction. Figure~\ref{fig:id} shows that our R-net trained with only image-level losses generates smoother textures and lower photometric errors than the compared opponents, but the resultant 3D shapes are less accurate by visual inspection.

To tackle this issue, we introduce a perception-level loss to further guide the training. Inspired by \cite{genova2018unsupervised}, we seek for the weak-supervision signal from a pre-trained deep face recognition network. Specifically, we extract the deep features of the images and compute the cosine distance:
\begin{equation}
\begin{split}
L_{per}(\bx) =  1 - \frac{<\!f(I),f(I'(\bx))>}{\|f(I)\|\cdot \|f(I'(\bx))\|}
\end{split}
\end{equation}
where $f(\cdot)$ denotes deep feature encoding and $<\!\cdot,\cdot\!>$ vector inner product. In this work, we train a FaceNet~\cite{schroff2015facenet} structure using an in-house face recognition dataset with 3M face images of 50K identities crawled from the Internet, and use it as our deep feature extractor.

Figure~\ref{fig:id} shows that with the perceptual loss, the textures are sharper and the shapes are more faithful. Quantitative results in the experiment section also show the benefit.

\subsection{Regularization}
To prevent face shape and texture degeneration, we add a commonly-used loss on the regressed 3DMM coefficients:
\begin{equation}
\begin{split}
L_{coef}(\bx) =  \omega_{\alpha}\|\balpha\|^2 + \omega_{\beta}\|\bbeta\|^2 + \omega_{\gamma}\|\bdelta\|^2
\end{split}
\end{equation}
which enforces a prior distribution towards the mean face. The balancing weights are empirically set to $\omega_{\alpha} = 1.0$,  $\omega_{\beta} = 0.8$ and $\omega_{\gamma} = 1.7e{-3}$.

Although the face textures in the Basel 2009 3DMM \cite{paysan20093d} were obtained with special devices, they still contain some baked-in shading (\eg, ambient occlusion). To favor a constant skin albedo similar to \cite{tewari2018self}, we add a flattening constrain to penalize the texture map variance:
\begin{equation}
\textstyle
L_{tex}(\bx)=\sum_{c \in \{r,g,b\}}var(\bT_{c,\mathcal{R}}(\bx))
\end{equation}

where $\mathcal{R}$ is a pre-defined skin region covering cheek, noise, and forehead. %(see \emph{suppl. material} for more details)

In summary, our loss function $L(\bx)$ for R-Net is composed of two image-level losses, a perceptual loss and two regularization loss. Their weights are set to $w_{photo}\!=\!1.9, w_{lan}\!=\!1.6e{-3}, w_{per}\!=\!0.2, w_{coef}\!=\!3e{-4}$ and $w_{tex}\!=\!5$ respectively in all our experiments.

\section{Weakly-supervised Neural Aggregation for Multi-Image Reconstruction}

Given multiple face images of a subject (\eg, a photo album), it is natural to leverage all the images to build a better 3D face shape. Images captured under different conditions should contain information complementary to each other due to change of pose, lighting \etc. Moreover, using an image set for reconstruction can gain further robustness to occlusion and bad lighting in some individual images.

Applying deep neural networks on an arbitrary number of orderless images is not straightforward. In this work, we use a network to learn a measurement of confidence or quality of the single-image reconstruction results, and use it to aggregate the individual shapes. Specifically, we seek to generate a vector $\bc \in \mathbb{R}^{80}$ with positive elements measuring the confidence of the identity-bearing shape coefficients $\balpha \in \mathbb{R}^{80}$. We do not consider other coefficients such as expression, pose, and lighting as they vary across images and fusion is unnecessary. We also bypass texture as we found the skin color of a subject can vary significantly across in-the-wild images. Let $\mathcal{I}:= \{I^j|j=1,\ldots,M\}$ be an image collection of a person, $\bx^j = (\balpha^j,\bbeta^j,\bdelta^j,\bp^j,\bgamma^j)$ the output coefficient vector from R-Net for each image $j$, and $\bc^j$ the confidence vector for each $\balpha^j$, we obtain the final shape via element-wise shape coefficient aggregation:
\begin{equation}
\textstyle
\balpha_{aggr} = ({\sum_{j} \bc^j\odot\balpha^j})~\oslash~({\sum_{j} \bc^j})\label{eq:aggr}
\end{equation}
where $\odot$ and $\oslash$ denote Hadamard product and division, respectively.

Next, we present how we train a network, denoted as C-Net, to predict $\bc$ in a weakly-supervised fashion without labels. The structure of C-Net will be presented afterwards.

\subsection{Label-Free Training}\label{section:C-Net}
To train C-Net on image sets, we generate the reconstructed image set $\{I^{j}{'}\}$ of $\{I^j\}$ with $\{\hat{\bx}^j\}$, where $\hat{\bx}^j = (\balpha_{aggr},\bbeta^j,\bdelta^j,\bp^j,\bgamma^j)$. We define the training loss as
\vspace{-6pt}
\begin{equation}
\mathcal{L}(\{\hat{\bx}^j\})=\frac{1}{M}\sum_{j=1}^M L(\hat{\bx}^j)
\vspace{-5pt}
\end{equation}
where $L(\cdot)$ is our hybrid-level loss function defined in Section~\ref{sec:hyrid} evaluated with $I^{j}{'}$ of $I^j$.

This way, the error can be backpropagated to $\balpha_{aggr}$ thus further to $\bc$ and C-Net weights, since Eq.~\ref{eq:aggr} is differentiable. C-Net will be trained to produce confidences that lead to an aggregated 3D face shape consistent with the face image set as much as possible. The pipeline is illustrated in Fig.~\ref{fig:overview}(c). In the multi-image training stage, the loss weights $\omega_{lm},\omega_{photo}$ and $\omega_{id}$ are set to $1.6e{-3}, 1.9$, and $0.1$ respectively.

Our aggregation design and training scheme are inspired by the set-based face recognition work of \cite{yang2017neural}. However, \cite{yang2017neural} used a scalar quality score for feature vector aggregation, whereas we produce element-wise scores for 3DMM coefficients. In Section~\ref{sec:multiimageabaltion}, we show element-wise scores yield superior results and analyze how our network exploits face pose difference for better shape aggregation.

\subsection{Confidence-Net Structure}
Our C-Net is designed to be light-weight. Since R-Net is able to predict high-level information such as pose and lighting, it is natural to reuse its feature maps for C-Net. In practice, we take both shallow and deep features from R-Net, as illustrated in Fig.~\ref{fig:overview} (a). The shallow feature can be used to measure image corruptions such as occlusion.

Specifically, we take the features after the first residual block $F_{b1} \in \mathbb{R}^{28\times28\times256}$ and after global pooling $F_{g} \in \mathbb{R}^{2048}$ of R-Net as the input to C-Net. We apply three $3\times3$ convolution layers $256$ channels and stride $2$, followed by a global pooling layer on $F_{b1}$ to get $F_{b1}' \in \mathbb{R}^{256}$. We then concatenate $F_{b1}'$ and $F_{g}$, and apply two fully-connected layers with 512 and 80 neurons respectively. At last, we apply sigmoid function to make the confidence predictions $\bc\in\mathbb{R}^{80}$ positive. Our C-Net has $3$M parameters in total, which is about $1/8$ size of R-Net.

\section{Experiments}
\noindent\textbf{Implementation Details.~}
To train our R-Net, we collected in-the-wild images from multiple sources such as CelebA~\cite{liu2015deep}, 300W-LP~\cite{zhu2016face}, I-JBA~\cite{klare2015pushing}, LFW~\cite{huang2007labeled} and LS3D~\cite{bulat2017far}. We balanced the pose and race distributions and get $\sim$260K face images as our training set. We use the method of \cite{chen2016supervised} to detect and align the images. The input image size is 224$\times$224. We take the weights pre-trained in ImageNet~\cite{russakovsky2015imagenet} as initialization, and train R-Net using Adam optimizer~\cite{kingma2015adam} with batch size of $5$, initial learning rate of $1e{-4}$, and $500$K total iterations.

To train C-Net, we construct an image corpus using 300W-LP~\cite{zhu2016face}, Multi-PIE~\cite{gross2010multi} and part of our in-house face recognition dataset.
For 300W-LP and Multi-PIE, we choose 5 images with rotation angles evenly distributed for each person. For the face recognition dataset, we randomly select 5 images for each person. The whole training set contains $\sim$50K images of $\sim$10K identities.
We freeze the trained R-Net, and randomly initialize C-Net except for its last fully-connected layer which is initialized to zero (so that we start from average pooling). We train it using Adam~\cite{kingma2015adam} with batch size of $5$, initial learning rate of $2e{-5}$ and $10$K total iterations.

\begin{table}[t]
	\centering
	\small
	\caption{Average reconstruction errors (mm) on MICC~\cite{bagdanov2011florence} and FaceWarehouse~\cite{cao2014facewarehouse} datasets for R-Net trained with different losses. Our full hybrid-level loss function yields significantly higher accuracy than other baselines on both datasets.}\label{table:loss}
	\begin{tabular}{ccccc}
		\hline
		\multicolumn{3}{c}{Losses} & \multirow{2}{*}{MICC}& \multirow{2}{*}{Facewarehouse}\\
		$L_{photo}$& $L_{lan}$& $L_{per}$& &\\
		\hline
		&&\checkmark&1.87$\pm$0.43&2.70$\pm$0.73\\
		\checkmark&\checkmark&& 1.80$\pm$0.52 &2.17$\pm$0.65\\
		&\checkmark&\checkmark &1.71$\pm$0.43 &2.11$\pm$0.48\\
		\checkmark&\checkmark&\checkmark& \textbf{1.67$\pm$0.50}& \textbf{1.81$\pm$0.50}\\
		\hline
	\end{tabular}
\end{table}

\subsection{Results on Single-Image Reconstruction}
\subsubsection{Ablation Study}
To validate the efficacy of our proposed hybrid-level loss function, we conduct ablation study on two datasets: the MICC Florence 3D Face dataset~\cite{bagdanov2011florence} and the  FaceWarehouse dataset~\cite{cao2014facewarehouse}. MICC contains 53 subjects, each associated with a ground truth scan in neutral expression and three video sequences captured in cooperative, indoor, and outdoor scenarios. For FaceWarehouse, we use 9 subjects each with 20 expressions for evaluation.

Table~\ref{table:loss} presents the reconstruction errors with various loss combinations. It shows that jointly considering image- and perception-level information gives rise to significantly higher accuracy than using them separately.

\vspace{-4pt}
\subsubsection{Comparison with Prior Art}\label{section:single}

\begin{table}[t]
	\centering
	\small
	\caption{Mean Root Mean Squared Error (RMSE) across 53 subjects on MICC dataset (in mm). We use ICP for alignment and compute point-to-plane distance between results and ground truth.\label{tab:micc}}
	\begin{tabular}{cccc}
		\hline
		Method & \!\!\!Cooperative\!\!\! & Indoor & Outdoor \\
		\hline
		\!\!Tran \etal~\cite{tran2017regressing}\!\! & \!1.97$\pm$0.49\! & \!2.03$\pm$0.45\! & \!1.93$\pm$0.49\!\\
		\!\!Genova \etal~\cite{genova2018unsupervised}\!\! & \!1.78$\pm$0.54\! & \!1.78$\pm$0.52\! & \!1.76$\pm$0.54\!\\
		Ours & \!\textbf{1.66$\pm$0.52}\! & \!\textbf{1.66$\pm$0.46}\! & \!\textbf{1.69$\pm$0.53}\! \\
		\hline
	\end{tabular}
	\vspace{-1pt}
\end{table}

\begin{figure}[t]
	\includegraphics[width=1.0\columnwidth]{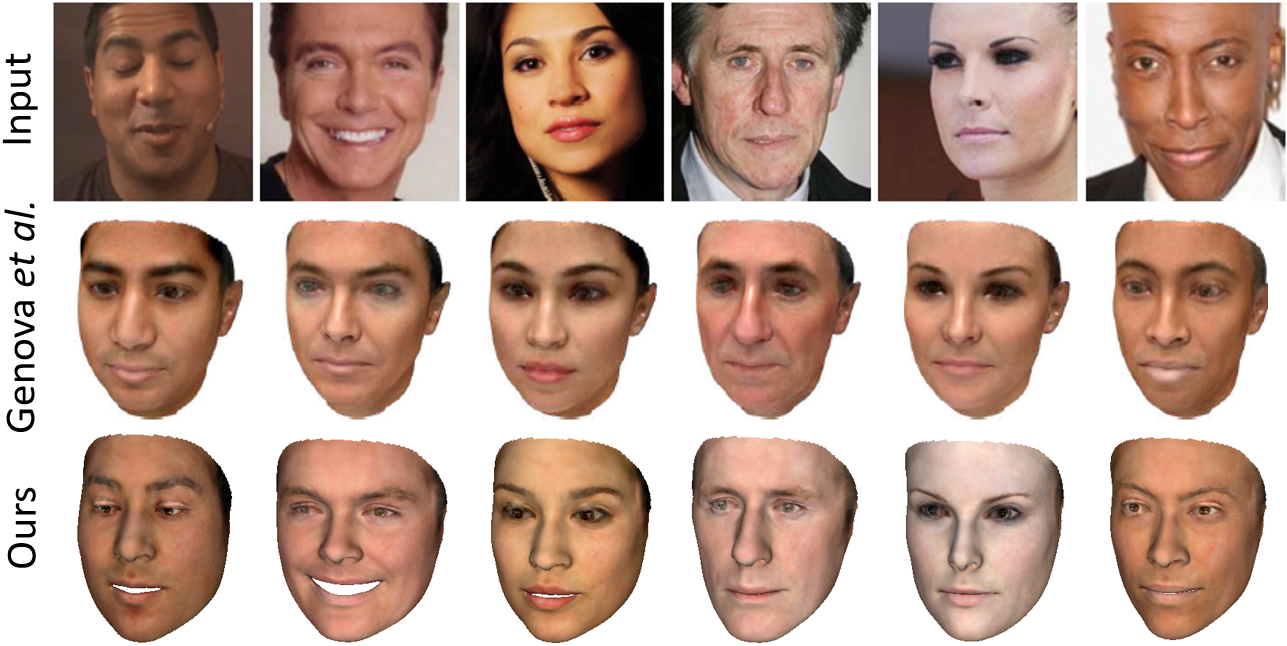}
	\caption{Comparison with Genova \etal~\cite{genova2018unsupervised}. Our texture and shape exhibit larger variance and are more consistent with the inputs. The images are from \cite{genova2018unsupervised}.}\label{quality_unmm}
\end{figure}

\begin{figure}[t]
	\includegraphics[width=1.0\columnwidth]{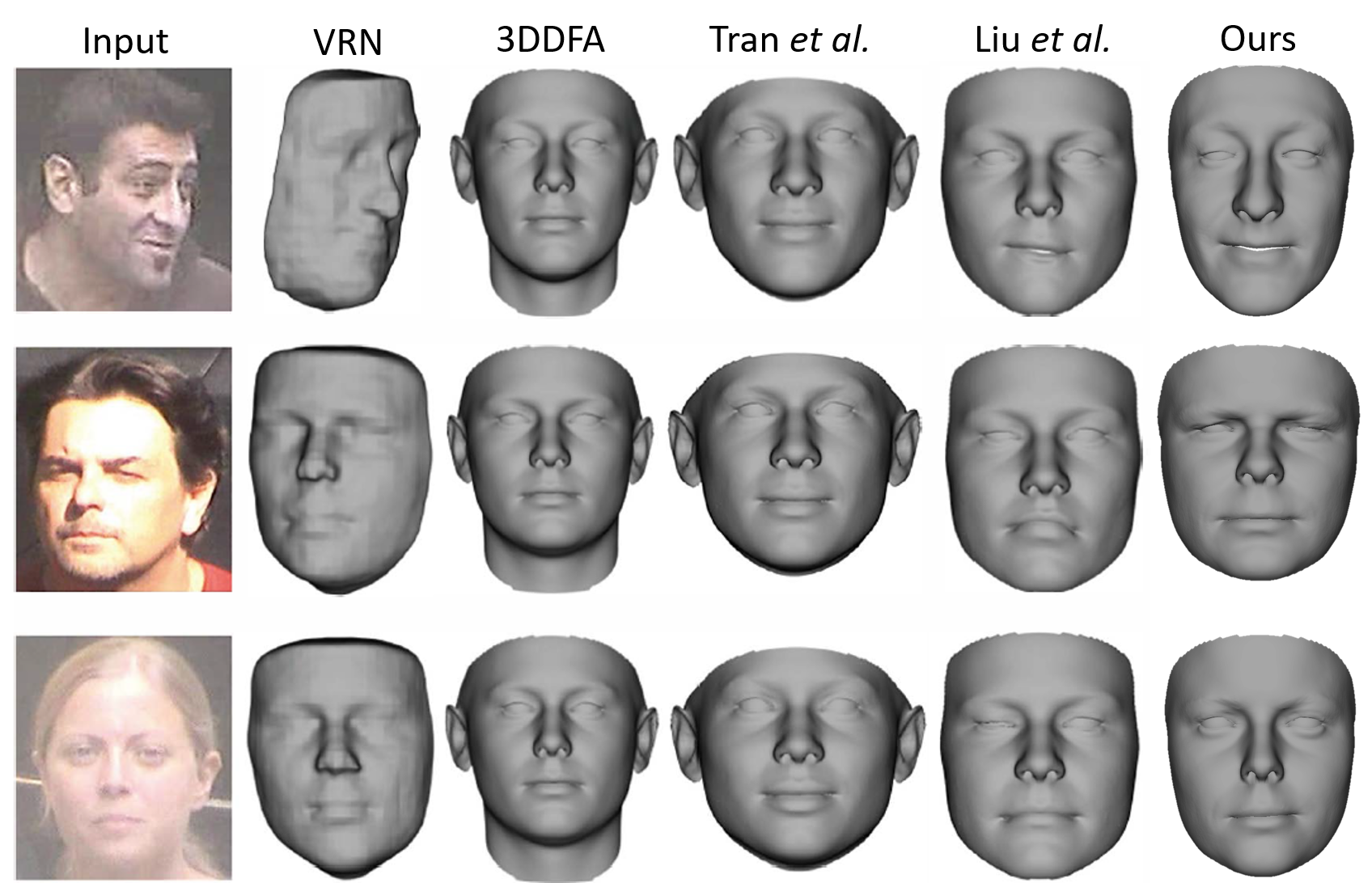}
	\caption{Comparison with VRN~\cite{jackson2017large}, 3DDFA~\cite{zhu2016face}, Tran \etal~\cite{tran2017regressing}, Liu \etal~\cite{liu2018disentangling} on three MICC subjects. Our results show largest variance and are visually most faithful among all methods. The input images and results of other methods are from \cite{liu2018disentangling}.}\label{fig:micc}
\end{figure}

\noindent\textbf{Comparison on MICC Florence with \cite{tran2017regressing,genova2018unsupervised,feng2018joint,jackson2017large,zhu2016face,liu2018disentangling}.~}
We first compare with the methods of Tran \etal~\cite{tran2017regressing} and Genova \etal~\cite{genova2018unsupervised}.
For \cite{tran2017regressing} and ours, we evaluate the error with the average shape from a sequence. \cite{genova2018unsupervised} averaged their encoder embeddings from all frames before reconstruction and produce a single shape per sequence. Following \cite{genova2018unsupervised}, we crop the ground truth mesh to 95mm around the nose tip and run ICP with isotropic scale for alignment. The results of \cite{tran2017regressing} only contains part of the forehead region, thus we further cut the ground truth meshes accordingly for fair comparison. Table~\ref{tab:micc} shows that our method significantly outperforms \cite{tran2017regressing} and \cite{genova2018unsupervised} on all three sequences. The qualitative comparison in Fig.~\ref{quality_unmm} and Fig.~\ref{fig:micc} also demonstrates the superiority of our results. Note that \cite{genova2018unsupervised} uses a perceptual loss similar to ours, but they ignores the low-level information such as photometric similarity. 


\begin{figure}[t]
	\centering
	\includegraphics[width=1.0\columnwidth]{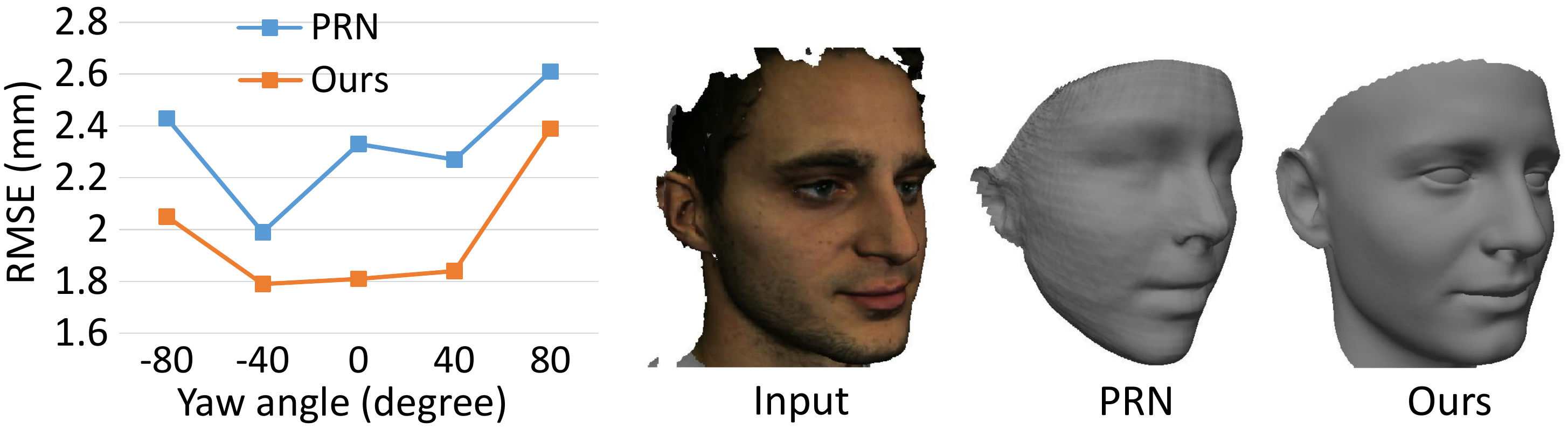}
	\caption{Comparison with PRN~\cite{feng2018joint} on MICC. Leftmost: Mean RMSE of different yaw angles. Our method excels at all views. Right three images: qualitative result comparison.}\label{fig:prn}
	\vspace{-2pt}
\end{figure}

We then compare with PRN~\cite{feng2018joint}, a recent CNN method with supervised learning that predicts unrestricted face shapes.
Following \cite{feng2018joint}, we render face images with 20 poses for each subject using pitch angles of $-15$, $20$, and $25$ degrees and yaw angles of $-80$, $40$, $0$, $40$, $80$ degrees. Figure~\ref{fig:prn} shows the point-to-plane RMSE averaged across subjects and pitch angles. Our method has a much lower error than PRN for all yaw angles. Also note that PRN has a larger model size than ours (160MB vs. 92MB).

We further qualitatively compare with several learning-based methods including VRN ~\cite{jackson2017large}, 3DDFA~\cite{zhu2016face}, and Liu \etal~\cite{liu2018disentangling}. Figure~\ref{fig:micc} shows that our method can well-recover both identity and expression, whereas the results of other methods have very low shape variance.

\vspace{6pt}
\noindent\textbf{Comparison on Facewarehouse with \cite{tewari2018self,tewari2017mofa,kim2018inversefacenet,garrido2016reconstruction}.}
We compare our results on the 9 Facewarehouse subjects selected by \cite{tewari2018self}, with three learning-based approaches
of Tewari \etal~\cite{tewari2017mofa,tewari2018self}, Kim \etal~\cite{kim2018inversefacenet} and an optimization-based approach of Garrido \etal~\cite{garrido2016reconstruction}.
The evaluation protocol of \cite{tewari2018self} is used.
%(see \emph{suppl. materal} for details). 

We evaluate two face regions: a smaller one same as
\setlength{\columnsep}{0pt}%
\setlength{\intextsep}{0pt}
\begin{wrapfigure}[6]{r}{2.1cm}
	\centering
	\includegraphics[width=1.5cm]{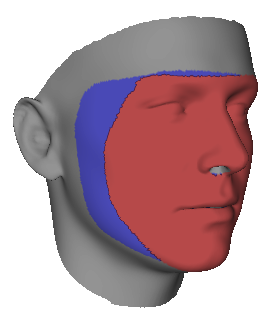}
	\caption{\label{fig:fwh_regions}}
\end{wrapfigure}
in \cite{tewari2018self}, and a larger one with more cheek areas included (see Fig.~\ref{fig:fwh_regions}). The point-to-point errors are presented in Table~\ref{table:faceware1}. Our method achieved the lowest reconstruction error among all learning-based methods. Note that \cite{tewari2017mofa}, \cite{tewari2018self}-C (coarse results), \cite{kim2018inversefacenet}, and our method are all based on 3DMM representation, and we show significant improvement upon theirs. Our method is even better than \cite{tewari2018self}-F which uses a corrective space to refine the 3DMM shape. Our accuracy gets closer to the optimization-based approach of \cite{garrido2016reconstruction} while our method can be orders of magnitude faster. 

\begin{figure}
	\includegraphics[width=1.0\columnwidth]{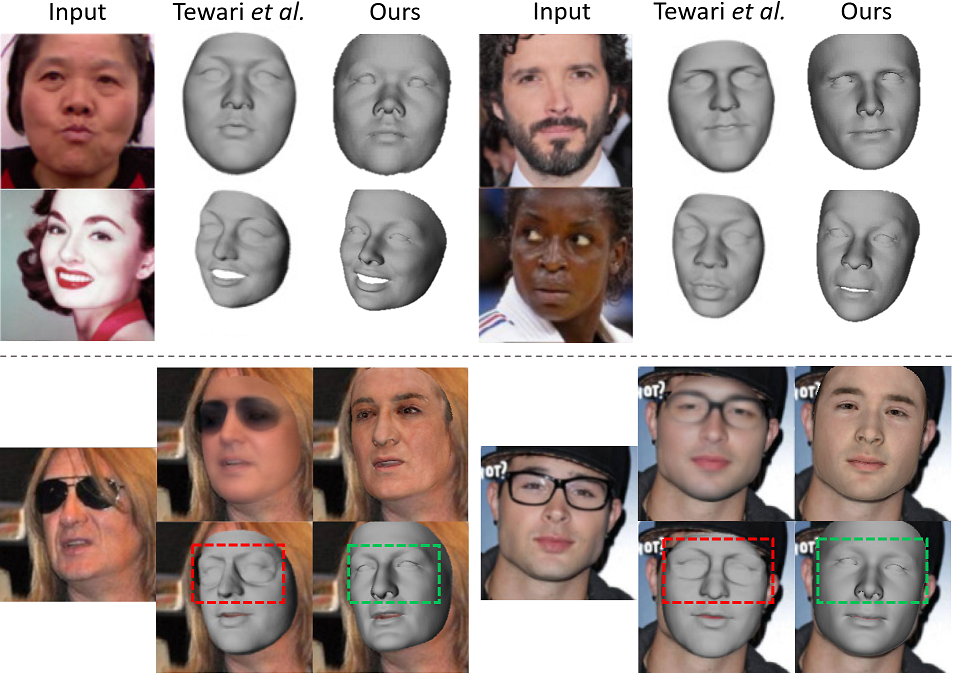}
	\caption{Comprison with Tewari \etal~\cite{tewari2018self} (fine results). Top: results on different races. Bottom: results under occlusion. The images are from \cite{tewari2018self}.
	}\label{fig:quality_250hz}
	\vspace{-2pt}
\end{figure}

\begin{figure}
	\centering
	\includegraphics[width=0.645\columnwidth]{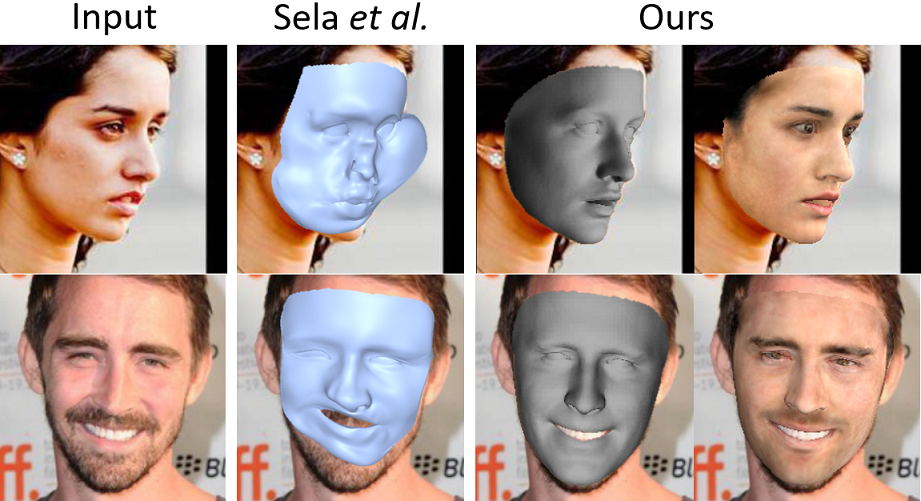}
	\caption{Comparison with Sela \etal~\cite{sela2017unrestricted} under large pose and challenging appearance.}\label{fig:quality_pix2v}
	\vspace{-1pt}
\end{figure}

We further compare with \cite{tewari2018self} qualitatively in Fig~\ref{fig:quality_250hz}. Our recovered shapes are of higher fidelity. Moreover, some artifacts from \cite{tewari2018self} can be observed under occlusion while our results are much more pleasing.
Also note that our method can handle profile faces (see, \eg, Fig.~\ref{fig:quality_pix2v}), while the large-pose robustness of the above methods are unclear to us.

\begin{table}[t]
	\centering
	\caption{Mean reconstruction error (mm) on 180 meshes of 9 subjects from FaceWarehouse. ``-F" and ``-C" denote the ``fine" and ``coarse" results of \cite{tewari2018self}. The face regions ``S" (Smaller) and ``L" (Larger) are shown in Fig.~\ref{fig:fwh_regions}. Our error is lowest among the learning-based methods. *: due to the GPU parallel computing scheme, one forward pass of our R-Net takes 20ms with both batch-size 1 and batch-size 10 (evaluated with an NVIDIA TITAN Xp GPU). The times of other methods are quoted from \cite{tewari2018self}.}\label{table:faceware1}
	\small
	\begin{tabular}{c|ccccc|c}
		\hline
		& \multicolumn{5}{c|}{Learning} & \!\!Optimization\!\!\!\\
		\cline{2-7}
		&\!\!Ours\!\!& \!\!\!\cite{tewari2018self}-F\!\!\!& \!\!\cite{tewari2018self}-C\!\!\!&\!\cite{tewari2017mofa}\!&\!\cite{kim2018inversefacenet}\!\!&\!\!\cite{garrido2016reconstruction}\!\!\!\\
		\hline
		\!\!Region-S\!\! & \textbf{1.81} & \!1.84\! &\!2.03\! &\!2.19\! &\!2.11\!\! & \!1.59\!\\
		\hline
		\!\!Region-L\!\! & \textbf{1.91} & \!2.00\! & - & - & -\! & \!1.84\!\\
		\hline
		$\stackrel{\hbox{Time}}{\hbox{~}}$ & \!\!{\small $\stackrel{\hbox{\!\!\!20ms}}{\hbox{(2ms}^*\hbox{)}}$}\!\!\! & \!$\stackrel{\hbox{4ms}}{\hbox{~}}$\! & \!$\stackrel{\hbox{4ms}}{\hbox{~}}$\! & \!$\stackrel{\hbox{4ms}}{\hbox{~}}$\! & \!$\stackrel{\hbox{4ms}}{\hbox{~}}$\!\! & \!$\stackrel{\hbox{120s}}{\hbox{~}}$\! \\
		\hline
	\end{tabular}
\end{table}

\begin{table}[t]
	\centering
	\caption{RMSE error (mm) of 100 faces in the BU-3DFE dataset.}\label{tab:bu3dfe}
	\small
	\begin{tabular}{c|ccc}
		\hline
		& Ours & PRN~\cite{feng2018joint} & Sela \etal~\cite{sela2017unrestricted} \\
		\hline
		Error& \!\textbf{1.40$\pm$0.31}\! & 1.86$\pm$0.47 & 2.91$\pm$0.60\\
		
		\hline
	\end{tabular}
\end{table}

\vspace{6pt}
\noindent\textbf{Comparison on BU-3DFE with \cite{sela2017unrestricted,feng2018joint}.~} 
The BU-3DFE dataset~\cite{yin20063d} contains 100 subjects across different races and each with different expressions. Here we use the frontal images of 100 neutral faces for our evaluation. Again, we perform ICP alignment and compute point-to-plane distance. Table~\ref{tab:bu3dfe} shows that our method significantly outperforms \cite{sela2017unrestricted} and \cite{feng2018joint}. The BU-3DFE images are captured under controlled settings. Figure~\ref{fig:quality_pix2v} further shows that our method also outperforms \cite{sela2017unrestricted} under large pose and challenging appearance variance of in-the-wild images.

\begin{figure}
	\centering
	\includegraphics[width=1.0\columnwidth]{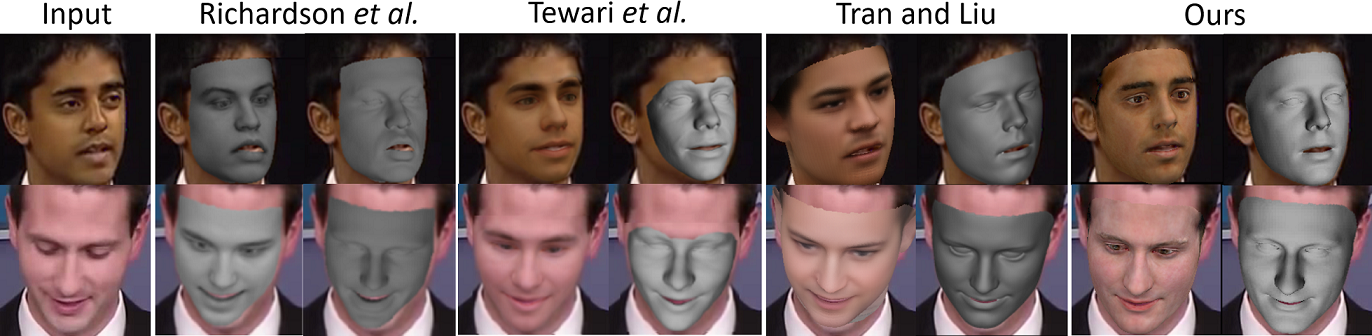}
	\caption{Comparison with Richardson \etal~\cite{richardson20163d}, Tewari \etal~\cite{tewari2017mofa}, and Tran and Liu~\cite{tran2018nonlinear}. Images are from \cite{tran2018nonlinear}. (Best viewed with zoom)}\label{fig:quality_nonlinear}
	\vspace{-2pt}
\end{figure}

\vspace{6pt}
\noindent\textbf{Comparison with other methods~\cite{richardson20163d,tran2018nonlinear}.~}
Figure~\ref{fig:quality_nonlinear} compares our results with Richardson \etal~\cite{richardson20163d}, Tran and Liu~\cite{tran2018nonlinear} and Tewari \etal~\cite{tewari2017mofa}. By visual inspection, our method produces better results.

\subsection{Results on Multi-Image Reconstruction}

\subsubsection{Ablation Study and Analysis}\label{sec:multiimageabaltion}
To test our multi-image shape aggregation method, we first conduct ablation study on render images of MICC. We render 20 poses for each of the 53 subjects as in Sec.~\ref{section:single}. Table~\ref{table:florence} presents the shape error of different aggregation strategies (S1 to S5). For S1, we train a C-Net similar to that described in Sec.~\ref{section:C-Net} but modify the final FC layer to output a global confidence score $\texttt{c}^j \in \mathbb{R^+}$, and we aggregate the identity coefficients via $\balpha_{aggr}=\sum_j\texttt{c}^j\cdot\balpha^j/\sum_j\texttt{c}^j$. For S2, we sum all elements in the confidence vector $\bc^j \in \mathbb{R}^{80}$ to get a global confidence for aggregation. For S3, we simply choose a single shape with largest confidence vector summation for error computation. For S4, we use our element-wise coefficient aggregation described in Sec.~\ref{section:C-Net}.

Table~\ref{table:florence} shows that all shape aggregation methods including the naive shape averaging have a lower error than the mean of per-frame shape errors. Nevertheless, all our aggregation strategies yield better results than naive shape averaging, demonstrating the efficacy of our learning-based aggregation method. Among them, the element-wise coefficient aggregation (S4) performs best.

\begin{table}[t]
	\centering
	\caption{Multi-image reconstruction errors on MICC rendered images with different aggregation strategies (see text for details).}\label{table:florence}
	\small
	\begin{tabular}{lc}
		\hline
		Shape error mean & 1.97$\pm$0.70 \\
		\hline
		Shape averaging  & 1.78$\pm$0.59 \\
		\hline
		Our S1: Global Aggr. with \texttt{c}$^j$  & 1.71$\pm$0.56\\
		Our S2: Global Aggr. with $\sum_i \bc^{j}_i$ & 1.70$\pm$0.55\\
		Our S3: Max Conf. $j=\argmax_j \sum \bc^{j}_i$ & 1.71$\pm$0.50\\
		Our S4: Elementwise Aggr. with $\bc^j$ & \textbf{1.67$\pm$0.54}\\
		\hline
	\end{tabular}
\end{table}

\begin{figure}
	\centering
	\includegraphics[width=1.0\columnwidth]{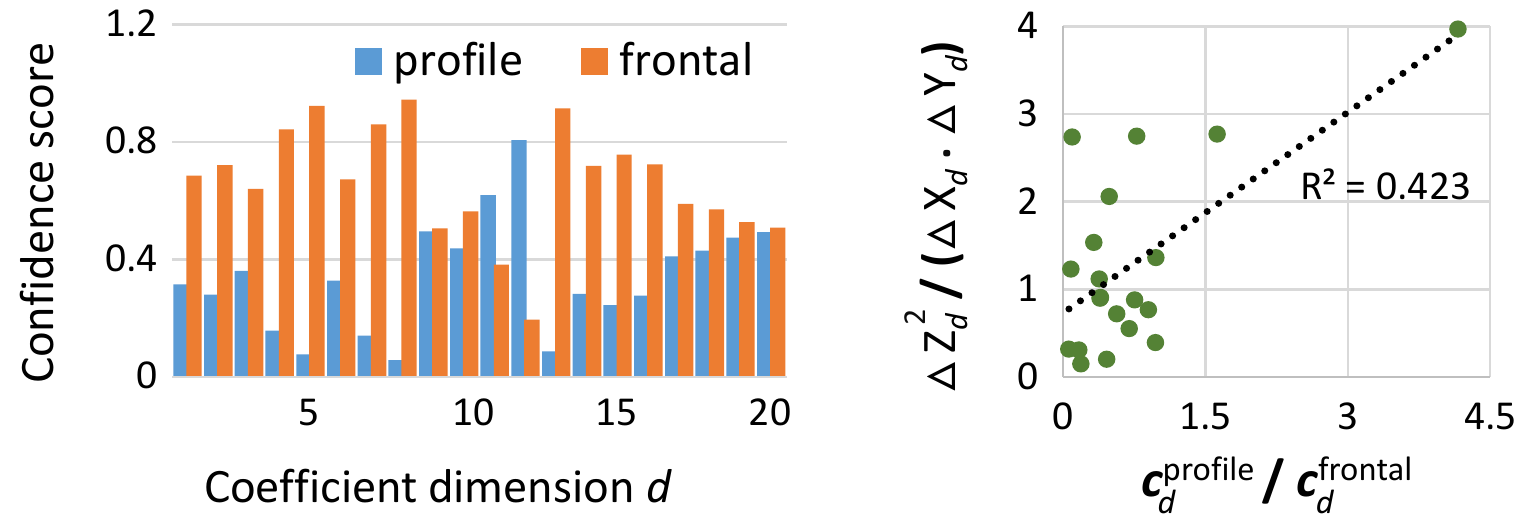}
	\caption{Confidence statistics on frontal and profile images. We show the first 20 entries with largest PCA energy (standard derivation). Left: average relative confidence scores of 53 subjects. Right: $Z$-direction shape influence \emph{w.r.t.} profile-to-frontal confidence ratio. Each dot represents a coefficient vector entry.
		Entries having larger influence on face depth ($Z$-direction) tend to get relatively larger confidence scores on profile faces than on frontal ones (linear regression $R^2 = 0.423$). }\label{fig:confidence}
\end{figure}
\begin{figure}
	\includegraphics[width=0.98\columnwidth]{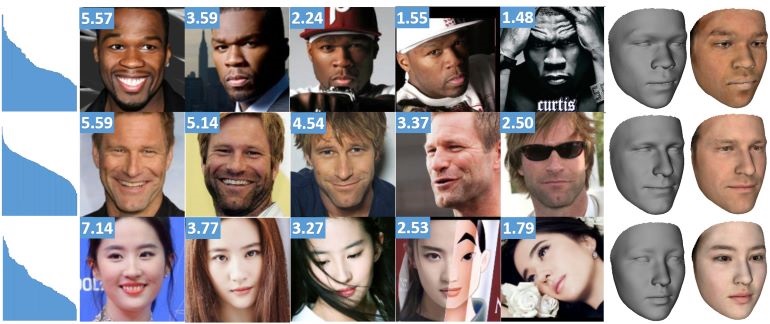}
	\caption{Results on in-the-wild image sets. The leftmost bar chart shows the sorted value of confidence vector summation of each image in the set. Five images sampled from a set are shown in the middle with their confidence vector summations shown in the top left corner. The last two columns are our final results.
	}\label{fig:wild}
\end{figure}

\vspace{5pt}
\noindent\textbf{Analysis.~}
We further analyze confidence score statistics to see whether they are affected by face pose. We compute the average relative confidence scores for profile and frontal images respectively. As shown in Fig~\ref{fig:confidence}~(left), for profile faces the confidence is lower in general, but higher on a few dimensions. Figure~\ref{fig:confidence}~(right) further shows that the coefficient entries having larger influence on face depth ($Z$-direction in the our 3D face coordinate system) tend to get relatively larger confidence scores on profile faces than on frontal ones. This is consistent with our intuition, and suggests that with element-wise confidences, the network can exploit view difference for better reconstruction. 
%Due to space constraint, more details about this experiment are deferred to the \emph{suppl. material}. 

Figure~\ref{fig:wild} presents some examples of our confidence prediction on our test set (to ease presentation we show the confidence vector summation $\sum_i\bc^j_i$ for each image). Our C-Net generally favors quality face images with frontal pose, high visibility, natural lighting etc. Occlusions like sun-glasses, hat and hair decrease the confidence.

\begin{table}[t]
	\centering
	\caption{Multi-image reconstruction error on the MICC dataset. We use same inputs and evaluation metric as in Table~\ref{tab:micc}. ``\cite{piotraschke2016automated}-G" and ``\cite{piotraschke2016automated}-S" denote global and segment-based aggregation of our predicted shapes using the strategy of \cite{piotraschke2016automated}.}\label{table:multi2}
	\small
	\begin{tabular}{ccccc}
		\hline
		Method & \!\!\!\!Cooperative\!\!\!\! & \!\!Indoor\!\! & \!\!Outdoor\!\! & \!\!All\!\!\\
		\hline
		\!\!\!Shape averaging\!\!\! & \!\!\!1.66$\pm$0.52\!\!\! & \!\!\!1.66$\pm$0.46\!\!\! & \!\!\!1.69$\pm$0.53\!\!\! & \!\!\!1.62$\pm$0.51\!\!\!\\
		\!\!\!\cite{piotraschke2016automated}-G\!\!\!& \!\!\!1.68$\pm$0.57\!\!\! & \!\!\!1.67$\pm$0.47\!\!\! & \!\!\!1.73$\pm$0.53\!\!\! & \!\!\!1.65$\pm$0.55\!\!\!\\
		\!\!\cite{piotraschke2016automated}-S\!\!& \!\!\!1.68$\pm$0.58\!\!\! & \!\!\!1.67$\pm$0.48\!\!\! & \!\!\!1.72$\pm$0.52\!\!\! & \!\!\!1.65$\pm$0.55\!\!\!\\
		\!\!Ours (S4)\!\! & \!\!\!\textbf{1.60$\pm$0.51}\!\!\! & \!\!\!\textbf{1.61$\pm$0.44}\!\!\! & \!\!\!\textbf{1.63$\pm$0.47}\!\!\! & \textbf{\!\!1.56$\pm$0.48\!\!\!}\\
		\hline
	\end{tabular}
	\vspace{-3pt}
\end{table}

\vspace{-3pt}
\subsubsection{Comparison with Prior Art}
\vspace{-1pt}
To our knowledge, our method is the first one applying neutral networks for face reconstruction confidence prediction and aggregation. So here we compare with a heuristic strategy of Piotraschke and Blanz~\cite{piotraschke2016automated}. Table~\ref{table:multi2} shows that our method produced better results than shape averaging and \cite{piotraschke2016automated} on the MICC dataset (we treat a sequence as an image set). The method of \cite{piotraschke2016automated} underperformed. Its results are even slightly worse than shape averaging. We conjecture this is because \cite{piotraschke2016automated} rely on the surface normal discrepancy with mean face to eliminate deficient reconstructions, yet our R-Net always produces a smooth, plausible face shape which renders their quality measurement ineffective.

\section{Conclusions}
We have proposed a CNN-based single-image face reconstruction method which exploits hybrid-level image information for weakly-supervised learning without ground-truth 3D shapes. Comprehensive experiments have shown that our method outperforms previous methods by a large margin in terms of both accuracy and robustness. We have also proposed a novel multi-image face reconstruction aggregation method using CNNs. Without any explicit label, our method can learn to measure image quality and exploit the complementary information in different images to reconstruct 3D faces more accurately.

{\small
	\bibliographystyle{ieee}
	\bibliography{facerecon}
}
	
\end{document}